\documentclass[runningheads]{llncs}
\usepackage{graphicx}
\usepackage{booktabs}
\usepackage{algorithm, algpseudocode}
\usepackage{subfig}
\usepackage{amsmath}
\usepackage{amssymb}
\usepackage{color, colortbl}
\usepackage{xcolor}
\usepackage{url}

\begin{document}
\title{Diverse Policies Converge in Reward-free Markov Decision Processes}
\titlerunning{Study on Diverse Policies}
%
\author{Fanqi Lin\inst{1} \and
Shiyu Huang\inst{2}\orcidID{0000-0003-0500-0141} \and
Wei-Wei Tu\inst{2}}
\authorrunning{Fanqi Lin et al.}
\institute{Tsinghua University, Beijing, 100084, China \\
\email{lfq20@mails.tsinghua.edu.cn}\\
4Paradigm Inc., Beijing, 100084, China\\
\email{\{huangshiyu,tuweiwei\}@4paradigm.com}}

\maketitle              
\begin{abstract}
Reinforcement learning has achieved great success in many decision-making tasks, and traditional reinforcement learning algorithms are mainly designed for obtaining a single optimal solution. However, recent works show the importance of developing diverse policies, which makes it an emerging research topic. Despite the variety of diversity reinforcement learning algorithms that have emerged, none of them theoretically answer the question of how the algorithm converges and how efficient the algorithm is. In this paper, we provide a unified diversity reinforcement learning framework and investigate the convergence of training diverse policies. Under such a framework, we also propose a provably efficient diversity reinforcement learning algorithm. Finally, we verify the effectiveness of our method through numerical experiments\footnote{Access the code on GitHub: \url{https://github.com/OpenRL-Lab/DiversePolicies}}.

\end{abstract}

\section{Introduction}

Reinforcement learning (RL) shows huge advantages in various decision-making tasks, such as recommendation systems~\cite{shi2019virtual,xue2022resact}, game AIs~\cite{berner2019dota,huang2021tikick} and robotic controls~\cite{yu2021learning,makoviychuk2021isaac}. While traditional RL algorithms can achieve superhuman performances on many public benchmarks, the obtained policy often falls into a fixed pattern. For example, previously trained agents may just overfit to a determined environment and could be vulnerable to environmental changes~\cite{ellis2022smacv2}. Finding diverse policies may increase the robustness of the agent~\cite{mahajan2019maven,kumar2020one}. Moreover, a fixed-pattern agent will easily be attacked~\cite{wang2022adversarial}, because the opponent can find its weakness with a series of attempts. If the agent could play the game with different strategies each round, it will be hard for the opponent to identify the upcoming strategy and it will be unable to apply corresponding attacking tactics~\cite{lanctot2017unified}. 
Recently, developing RL algorithms for diverse policies has attracted the attention of the RL community for the promising value of its application and also for the challenge of solving a more complex RL problem~\cite{eysenbach2018diversity,huang2022vmapd,chen2023dgpo}.

Current diversity RL algorithms vary widely due to factors like policy diversity measurement, optimization techniques, training strategies, and application scenarios. This variation makes comparison challenging. While these algorithms often incorporate deep neural networks and empirical tests for comparison, they typically lack in-depth theoretical analysis on training convergence and algorithm complexity, hindering the development of more efficient algorithms.

To address the aforementioned issues, we abstract various diversity RL algorithms, break down the training process, and introduce a unified framework. We offer a convergence analysis for policy population and utilize the contextual bandit formulation to design a more efficient diversity RL algorithm, analyzing its complexity. We conclude with visualizations, experimental evaluations, and an ablation study comparing training efficiencies of different methods.
We summarise our contributions as follows: (1) We investigate recent diversity reinforcement learning algorithms and propose a unified framework. (2) We give out the theoretical analysis of the convergence of the proposed framework. (3) We propose a provably efficient diversity reinforcement learning algorithm. (4) We conduct numerical experiments to verify the effectiveness of our method.

\section{Related Work}
\paragraph{{\bf Diversity Reinforcement Learning}}
Recently, many researchers are committed to the design of diversity reinforcement learning algorithms~\cite{eysenbach2018diversity,osa2022discovering,huang2022vmapd,chen2023dgpo}. DIYAN~\cite{eysenbach2018diversity} is a classical diversity RL algorithm, which learns maximum entropy policies via maximizing the mutual information between states and skills. Besides, \cite{osa2022discovering} trains agents with latent conditioned policies which make use of continuous low-dimensional latent variables, thus it can obtain infinite qualified solutions. 
More recently, RSPO~\cite{zhou2021continuously} obtains diverse behaviors via iteratively optimizing each policy. DGPO~\cite{chen2023dgpo} then proposes a more efficient diversity RL algorithm with a novel diversity reward via sharing parameters between policies.

\paragraph{{\bf Bandit Algorithms}}
The challenge in multi-armed bandit algorithm design is balancing exploration and exploitation. Building on $\epsilon$-greedy\cite{watkins1989learning}, UCB algorithms\cite{auer2002using} introduce guided exploration. Contextual bandit algorithms, like \cite{may2012optimistic,li2010contextual}, improve modeling for recommendation and reinforcement learning. They demonstrate better convergence properties with contextual information\cite{chu2011contextual,li2010contextual}. Extensive research\cite{auer2002finite} provides regret bounds for these algorithms.

\section{Preliminaries}

\paragraph{{\bf Markov Decision Process}}
We consider environments that can be represented as a Markov decision process (MDP). An MDP can be represented as a tuple $(\mathcal{S}, \mathcal{A}, P_T, r, \gamma)$, where $\mathcal{S}$ is the state space, $\mathcal{A}$ is the action space and $\gamma\in[0,1)$ is the reward discount factor. The state-transition function $P_T(s,a,s') : \mathcal{S} \times \mathcal{A} \times {S} \mapsto [0,1]$ defines the transition probability over the next state $s'$ after taking action $a$ at state $s$.
$r(s, a):\mathcal{S} \times \mathcal{A} \rightarrow \mathbb{R}$ is the reward function denoting the immediate reward received by the agent when taking action $a$ in state $s$. The discounted state occupancy measure of policy $\pi$ is denoted as $\rho^{\pi}(s)=(1-\gamma)\sum_{t=0}^{\infty}\gamma^tP_t^{\pi}(s)$, where $P_t^{\pi}(s)$ is the probability that policy $\pi$ visit state $s$ at time $t$. The agent' objective is to learn a policy $\pi$ to maximize the expected accumulated reward $J(\theta)=\mathbb{E}_{z\sim p(z),s\sim\rho^{\pi}(s), a\sim\pi(\cdot|s,z)}[\sum_t\gamma^t r(s_t,a_t)]$.
In diversity reinforcement learning, the latent conditioned policy is widely used. The latent conditioned policy is denoted as $\pi(a|s,z)$, and the latent conditioned critic network is denoted as $V^{\pi}(s,z)$. During execution, the latent variable $z\sim p(z)$ is randomly sampled at the beginning of each episode and keeps fixed for the entire episode. When the latent variable $z$ is discrete, it can be sampled from a categorical distribution with $N_z$ categories. When the latent variable $z$ is continuous, it can be sampled from a Gaussian distribution.


\begin{table*}[t]
\begin{center}
        \caption{Comparison of different diversity algorithms.}
        \label{tb:methodcompare}
        \begin{tabular}{ccccccc}
            \toprule
                Method & Citation & Policy Selection & Reward Calculation \\
            \midrule
            RSPO & \cite{zhou2021continuously} & Iteration Fashion & Behavior-driven / Reward-driven exploration \\
                SIPO & \cite{fu2022iteratively} & Iteration Fashion & Behavior-driven exploration \\
                DIAYN & \cite{eysenbach2018diversity} & Uniform Sample & $I(s;z)$ \\
                DSP & \cite{zahavy2021discovering} & Uniform Sample & $I(s, a; z)$ \\
            DGPO & \cite{chen2023dgpo} & Uniform Sample & $\min_{z'\ne z} D_{KL}(\rho^{\pi_{\theta}}(s|z) || \rho^{\pi_{\theta}}(s|z'))$ \\
            \midrule
            Our work & & Bandit Selection & Any form mentioned above \\
            \bottomrule
        \end{tabular}
    \end{center}

\end{table*}

\section{Methodology}
In this section, we will provide a theoretical analysis of diversity algorithms in detail. Firstly, in section~\ref{subsection: unified}, we propose a unified framework for diversity algorithms, and point out major differences between diversity algorithms in this unified framework. Then we prove the convergence of diversity algorithms in section~\ref{subsection: convergence}. We further formulate the diversity optimization problem as a contextual bandit problem, and propose {\bf bandit selection} in section~\ref{subsection: bandit}. Finally, we provide rigorous proof for $regret$ bound of {\bf bandit selection} in section~\ref{subsection: regret}.

\subsection{A Unified Framework for Diversity Algorithms}
\label{subsection: unified}
Although there has been a lot of work on exploring diversity, we find that these algorithms lack a unified framework. So we propose a unified framework for diversity algorithms in Algorithm~\ref{algo: unified framework} to pave the way for further research. 

We use $Div$ to measure the diversity distance between two policies and we abbreviate policy $\pi_\theta(\cdot|s,z_i)$ as $\pi^i$. Vector $z_i$ can be thought of as a skill unique to each policy $\pi^i$. Moreover, we define $U\in \mathbb{R}^{N \times N}$ as diversity matrix where $U_{ij} = Div(\pi^i, \pi^j)$ and $N$ denotes the number of policies. \\
For each episode, we first sample $z_i$ to decide which policy to update. Then we interact the chosen policy with the environment to get trajectory $\tau$, which is used to calculate intrinsic reward $r^{in}$ and update diversity matrix $U$. We then store tuple ($s, a,s',r^{in},z_i$) in replay buffer $\mathcal{D}$ and update $\pi^i$ through any reinforcement learning algorithm. \\
Here we abstract the procedure of selecting $z_i$ and calculating $r^{in}$ as $SelectZ$ and $CalR$ functions respectively, which are usually the most essential differences between diversity algorithms. We summarize the comparison of some diversity algorithms in Table~\ref{tb:methodcompare}. Now we describe these two functions in more detail. \\
{\bf Policy Selection.} Note that we denote by $p(z)$ the distribution of $z$. We can divide means to select $z_i$ into three categories in general, namely {\bf iteration fashion}, {\bf uniform sample} and {\bf bandit selection}: 

(1) {\bf Iteration fashion.}
Diversity algorithms such as RSPO~\cite{zhou2021continuously} and SIPO~\cite{fu2022iteratively} obtain diverse policies in an iterative manner. In the $k$-th iteration, policy $\pi^k$ will be chosen to update, and the target of optimization is to make $\pi^k$ sufficiently different from previously discovered policies $\pi^1, ..., \pi^{k-1}$. This method doesn't ensure optimal performance and is greatly affected by policy initialization.

(2) {\bf Uniform sample.}
Another kind of popular diversity algorithm such as DIAYN~\cite{eysenbach2018diversity} and DGPO~\cite{chen2023dgpo}, samples $z_i$ uniformly to maximize the entropy of $p(z)$. Due to the method's disregard for the differences between policies, it often leads to slower convergence.

(3) {\bf Bandit selection.}
We frame obtaining diverse policies as a contextual bandit problem. Sampling $z_i$ corresponds to minimizing regret in this context. This approach guarantees strong performance and rapid convergence.\\
{\bf Reward Calculation.} Diversity algorithms differ in intrinsic reward calculation. Some, like \cite{chen2023dgpo,eysenbach2018diversity,osa2022discovering}, use mutual information theory and a discriminator $\phi$ to distinguish policies. DIAYN\cite{eysenbach2018diversity} emphasizes deriving skill $z$ from the state $s$, while \cite{osa2022discovering} suggests using state-action pairs. On the other hand, algorithms like \cite{liu2021unifying,zhou2021continuously} aim to make policies' action or reward distributions distinguishable, known as behavior-driven and reward-driven exploration. DGPO\cite{chen2023dgpo} maximizes the minimal diversity distance between policies.

\begin{algorithm}[t]
\caption{A Unified Framework for Diversity Algorithms}
\label{algo: unified framework}
\begin{algorithmic}
\State {\bf Initialize:} $\pi_\theta(\cdot|s,z); U\in \mathbb{R}^{N \times N}(U_{ij} = Div(\pi ^ i , \pi^j)$) 
\For{each episode}
    \State Sample $z_i \sim SelectZ(U)$; 
    \State Get trajectory $\tau$ from $\pi^i$; 
    \State Get $r^{in}=CalR(\tau)$ and update $U$; 
    \State Store tuple ($s,a,s',r^{in},z_i$) in replay buffer $\mathcal{D}$; 
    \State Update $\pi^i$ with $\mathcal{D}$;
\EndFor
\end{algorithmic}
\end{algorithm}
\subsection{Convergence Analysis}
\label{subsection: convergence}
In this section, we will show the convergence of diversity algorithms under a reasonable diversity target. We define $\mathcal{P} = \{\pi^1, \pi^2, ..., \pi^N\}$ as the set of independent policies, or policy population.

{\bf Definition 1.} $g: \{\pi^1, \pi^2, ..., \pi^N\} \to \mathbb{R}^{N \times N}$ is a function that maps population $\mathcal{P}$ to diversity matrix $U$ which is defined in section~\ref{subsection: unified}. Given a population $\mathcal{P}$, we can calculate pairwise diversity distance under a certain diversity metric, which indicates that $g$ is an injective function. 

{\bf Definition 2.} Note that in the iterative process of the diversity algorithm, we update $\mathcal{P}$ directly instead of $U$. So if we find a valid $U$ that satisfies the diversity target, then the corresponding population $\mathcal{P}$ is exactly our target diverse population. We refer to this process of finding $\mathcal{P}$ backward as $g^{-1}$.

{\bf Definition 3.} $f: \mathbb{R}^{N \times N} \to R$ is a function that maps $U$ to a real number. While $U$ measures the pairwise diversity distance between policies, $f$ measures the diversity of the entire population $\mathcal{P}$. As the diversity of the population increases, the diversity metric calculated by $f$ will increase as well.

{\bf Definition 4.} We further define $\delta$\emph{-target population set} $\mathcal{T}_{\delta} = \{g^{-1}(U) | f(U) > \delta, U \in \mathbb{R}^{N \times N}\}$. $\delta$ is a threshold used to separate target and non-target regions. The meaning of this definition is that, during the training iteration process, when the diversity metric closely related to $U$ exceeds a certain threshold, or we say $f(U) > \delta$, the corresponding population $\mathcal{P}$ is our target population. \\
Note two important points: (1) The population meeting the diversity requirement should be a set, not a fixed point. (2) Choose a reasonable threshold $\delta$ that ensures both sufficient diversity and ease of obtaining the population.

\begin{theorem}
$(\frac{\partial f}{\partial U})_{ij} =  \frac{\partial f}{\partial U_{ij}} = \frac{\partial f}{\partial Div(\pi^i, \pi^j)} > 0$, where i, j $\in\{1, 2, 3, ..., N\}.$  \\
Proof. $f$ measures the diversity of the entire population $\mathcal{P}$. When the diversity distance between two policies in a population $\pi^i$ and $\pi^j$  increases, the overall diversity metric $f(U)$ will obviously increase. 
\end{theorem}

\begin{theorem}
\label{increase}
We can find some special continuous differentiable $f$ that, $\exists \varepsilon >0$, s.t. $(\frac{\partial f}{\partial U})_{ij} > \varepsilon$, where i, j $\in\{1, 2, 3, ..., N\}.$ \\
Proof. For example, we can simply define $f(U) = \sum_{i\ne j}U_{ij}$, where $(\frac{\partial f}{\partial U})_{ij} = 1$. So we can choose threshold $0<\varepsilon<1$, then we can find $(\frac{\partial f}{\partial U})_{ij} > \varepsilon$ obviously. Of course, we can also choose other relatively complex $f$ as the diversity metric.
\end{theorem}

\begin{theorem}
\label{diversity algo}
There's a diversity algorithm and a threshold $\nu>0$. Each time the population $\mathcal{P}$ is updated, several elements in U will increase by at least $\nu$ in terms of mathematical expectation. \\
Proof. In fact, many existing diversity algorithms already have this property. Suppose we currently choose $\pi^i$ to update. For DIAYN~\cite{eysenbach2018diversity}, $Div(\pi^i, \pi^j)$ and $Div(\pi^j, \pi^i)(\forall j \ne i)$ are increased in the optimization process. And for DGPO~\cite{chen2023dgpo}, suppose policy $\pi^j$ is the closest to policy $\pi^i$ in the policy space, then $Div(\pi^i, \pi^j)$ and $Div(\pi^j, \pi^i)$ are increased as well in the optimization process. Apart from these two, there are many other existing diversity algorithms such as ~\cite{osa2022discovering,zhou2021continuously,liu2021unifying} that share the same property. Note that we propose Theorem~\ref{diversity algo} from the perspective of mathematical expectation, so we can infer that, $\exists \nu > 0, j \ne i$, s.t. $Div(\pi'^i, \pi^j) - Div(\pi^i, \pi^j) > \nu$, where policy $\pi'^i$ denotes the updated policy $\pi^i$. And for $k \notin \{i,j\}$, we can assume $U_{ik}$ and $U_{ki}$ are unchanged for simplicity.
\end{theorem}

\begin{theorem}
With an effective diversity algorithm and a reasonable diversity $\delta$-target, we can obtain a diverse population $\mathcal{P} \in T_{\delta}$. \\
Proof. We denote by $\mathcal{P}_0$ the initialized policy population, and we define $f_0 = f(g(\mathcal{P}_0))$. Then $\exists M \in \mathcal{N}$, s.t. $f_0 + M\cdot\nu\varepsilon > \delta$. Given Theorem~\ref{increase} and Theorem~\ref{diversity algo}, we define $\mathcal{P}_M$ as the 
policy population after M iterations, then we have $f(g(\mathcal{P}_M)) > f_0 + M\cdot\nu\varepsilon$, which means we can obtain the $\delta$-target policy population in up to $M$ iterations. Or we can say that the diversity algorithm will converge after at most $M$ iterations.
\end{theorem}

{\bf Remark.} 
Careful selection of threshold $\delta$ is crucial for diversity algorithms. Reasonable diversity goals should be set to avoid difficulty or getting stuck in the training process. This hyperparameter can be obtained through empirical experiments or methods like hyperparameter search. In certain diversity algorithms, both $\delta$ and $\mathcal{P}$ may change during training. For instance, in {\bf iteration fashion} algorithms (Section~\ref{subsection: unified}), during the $k$-th iteration, $\mathcal{P} = \{\pi^1, \pi^2, ..., \pi^k\}$ with a target threshold of $\delta_k$. If policy $\pi^k$ becomes distinct from $\pi^1, ..., \pi^{k-1}$, meeting the diversity target, policy $\pi_{k+1}$ is added to $\mathcal{P}$ and the threshold changes to $\delta_{k+1}$.

\subsection{A Contextual Bandit Formulation}
\label{subsection: bandit}
As mentioned in Section~\ref{subsection: unified}, we can sample $z_i$ via {\bf bandit selection}. In this section, we formally define $K$-armed contextual bandit problem~\cite{li2010contextual}, and show how it models diversity optimization procedure. 

\begin{algorithm}[h!]
\caption{A Contextual Bandit Formulation}
\label{algo: Contextual Bandit}
\begin{algorithmic}
\State {\bf Initialize:} Arm Set $\mathcal{A}$; Contextual Bandit Algorithm $Algo$ 
\For{$t = 1, 2, 3, ...$}
    \State Observe feature vectors $x_{t,a}$ for each $a \in \mathcal{A}$; 
    \State Based on $\left \{x_{t,a}\right \}_{a \in \mathcal{A}}$ and reward in previous iterations, $Algo$ chooses an arm $a_{t} \in \mathcal{A}$ and receives reward $r_{t, a_t}$; 
    \State Update $Algo$ with ($x_{t,a_{t}}, a_{t}, r_{t, a_t}$);
\EndFor
\end{algorithmic}
\end{algorithm}

We show the procedure of the contextual bandit problem in Algorithm~\ref{algo: Contextual Bandit}. In each iteration, we can observe feature vectors  $x_{t, a}$ for each $a \in \mathcal{A}$, which are also denoted as \emph{context}. Note that \emph{context} may change during training. Then, $Algo$ will choose an arm $a_{t} \in \mathcal{A}$ based on contextual information and will receive reward $r_{t, a_t}$. Finally, tuple ($x_{t, a_{t}}, a_{t}, r_{t, a_t}$) will be used to update $Algo$. \\
We further define \emph{T-Reward}~\cite{li2010contextual} of $Algo$ as $\sum_{t=1}^{T}r_t$. Similarly, we define the \emph{optimal expected T-Reward} as ${\bf E}[\sum_{t=1}^{T}r_{t, a_{t}^*}]$, where $a_{t}^*$ denotes the arm with maximum expected reward in iteration $t$. To measure $Algo$'s performance, we define \emph{T-regret} $R_T$ of $Algo$ by 
\begin{equation}\label{eq1}
R_T = {\bf E}[\sum_{t=1}^{T}r_{t, a_{t}^*}] - {\bf E}[\sum_{t=1}^{T}r_{t, a_{t}}].
\end{equation}
Our goal is to minimize $R_T$.\\
In the diversity optimization problem, policies are akin to arms, and \emph{context} is represented by visited states or $\rho^{\pi}(s)$. Note that \emph{context} may change as policies evolve. When updating a policy, the reward is the difference in diversity metric before and after the update, linked to the diversity matrix $U$ (Section~\ref{subsection: unified}). Our objective is to maximize policy diversity, equivalent to maximizing expected reward or minimizing $R_T$ in contextual bandit formulation.\\
Here's an example to demonstrate the effectiveness of {\bf bandit selection}. In some cases, a policy $\pi^i$ may already be distinct enough from others, meaning that selecting $\pi^i$ for an update wouldn't significantly affect policy diversity. To address this, we should decrease the probability of sampling $\pi^i$. Fixed uniform sampling fails to address this issue, but bandit algorithms like UCB\cite{auer2002finite} or LinUCB\cite{li2010contextual} consider both historical rewards and the number of times policies have been chosen. This caters to our needs in such cases.

\subsection{Regret Bound}
\label{subsection: regret}
In this section, we provide the \emph{regret} bound for {\bf bandit selection} in the diversity algorithms.

{\bf Problem Setting.} We define $T$ as the number of iterations. In each iteration $t$, we can observe $N$ feature vectors $x_{t,a} \in \mathbb{R}^d$ and receive reward $r_{t, a_t}$ with $\Vert x_{t,a} \Vert \le 1$ for $a \in \mathcal{A}$ and $r_{t, a_t} \in [0, 1]$, where $\Vert \cdot \Vert$ means $l_2$-norm, $d$ denotes the dimension of feature vector and $a_t$ is the chosen action in iteration $t$.

{\bf Linear Realizability Assumption.} 
Similar to lots of theoretical analyses of contextual bandit problems~\cite{auer2002using,chu2011contextual}, we propose linear realizability assumption to simplify the problem. We assume that there exists an unknown weight vector $\theta^* \in \mathbb{R}^d$ with $\Vert \theta^* \Vert \le 1$ s.t. 
\begin{equation}\label{eq2}
{\bf E}[r_{t,a} | x_{t,a}] = x_{t,a}^T \theta^*.
\end{equation}
for all t and a. \\
We now analyze the rationality of this assumption in practical diversity algorithms. Reward $r_{t, a}$ measures the changed value of overall diversity metric $\bigtriangleup f(U)$ of policy population $\mathcal{P}$ after an update. Suppose $\pi^i_t$ is the policy corresponding to the feature vector $x_{t, a}$ in the iteration $t$. While $x_{t, a}$ encodes state features of $\pi^i_t$, it can encode the diversity information of $\pi^i_t$ as well. Therefore, we can conclude that $r_{t, a}$ is closely related to $x_{t, a}$. So given that $x_{t, a}$ contains enough diversity information, we can assume that the hypothesis holds.

\begin{theorem}
(Diversity Reinforcement Learning Oracle $\mathcal{DRLO}$). Given a reasonable $\delta$-target and an effective diversity algorithm, let the probability that the policy population $\mathcal{P}$ reaches $\delta$-target in $T$ iterations be $1-\backepsilon_{\delta, T}$. Then we have $\lim_{T \to \infty}\backepsilon_{\delta, T} = 0$. \\
Proof. This is actually another formal description of the convergence of diversity algorithms which has been proved in Section~\ref{subsection: convergence}. Experimental results~\cite{osa2022discovering,chen2023dgpo} have shown that $\backepsilon_{\delta, T}$ will decrease significantly when $T$ reaches a certain value.
\end{theorem}

\begin{theorem}
(Contextual Bandit Algorithm Oracle $\mathcal{CBAO}$). There exists a contextual bandit algorithm that makes regret bounded by $O\left (\sqrt{Td {\rm ln}^3 (NT{\rm ln}(T)/ \eta)} \right )$ for $T$ iterations with probability $1 - \eta$.\\
Proof. Different contextual bandit algorithm corresponds to different regret bound. In fact, we can use the regret bound of any contextual bandit algorithm here. The regret bound mentioned here is the regret bound of SupLinUCB algorithm~\cite{chu2011contextual}. For concrete proof of this regret bound, we refer the reader to ~\cite{chu2011contextual}.
\end{theorem}

\begin{theorem}
\label{bound}
For T iterations, the regret for {\bf bandit selection} in diversity algorithms is bounded by $O\left (\sqrt{Td {\rm ln}^3 (\frac{NT{\rm ln}(T)(1 - \backepsilon_{\delta, T})}{\eta - \backepsilon_{\delta, T}})} \right )$ with probability $1 - \eta$. Note that $\lim_{T \to \infty}(\eta - \backepsilon_{\delta, T}) = \eta > 0$. \\
Proof. In diversity algorithms, the calculation of the regret bound is based on the premise that a certain $\delta$-target has been achieved. Note that $\mathcal{DRLO}$ and $\mathcal{CBAO}$ are independent variables in this problem setting. Given $0<\eta<1$, we define
\begin{equation}\label{eq3}
\eta_1 = \frac{\eta - \backepsilon_{\delta, T}}{1 - \backepsilon_{\delta, T}}.
\end{equation}
Then we have 
\begin{equation}\label{eq4}
1 - \eta = (1 - \backepsilon_{\delta, T})(1 - \eta_1).
\end{equation}
The implication of Equation~\ref{eq4} is that, for $T$ iterations, with probability $1 - \eta$, the regret for {\bf bandit selection} in diversity algorithms is bounded by
\begin{equation}\label{eq5}
\begin{split}
O\left (\sqrt{Td {\rm ln}^3 (NT{\rm ln}(T)/ \eta_1)} \right )
= O\left (\sqrt{Td {\rm ln}^3 (\frac{NT{\rm ln}(T)(1 - \backepsilon_{\delta, T})}{\eta - \backepsilon_{\delta, T}})} \right ).
\end{split}
\end{equation}
The right-hand side of Equation~\ref{eq5} is exactly the regret bound we propose in Theorem~\ref{bound}.
\end{theorem}

\begin{figure}[t]
\center
\subfloat[]{\begin{centering}
\includegraphics[width=0.2375\linewidth]{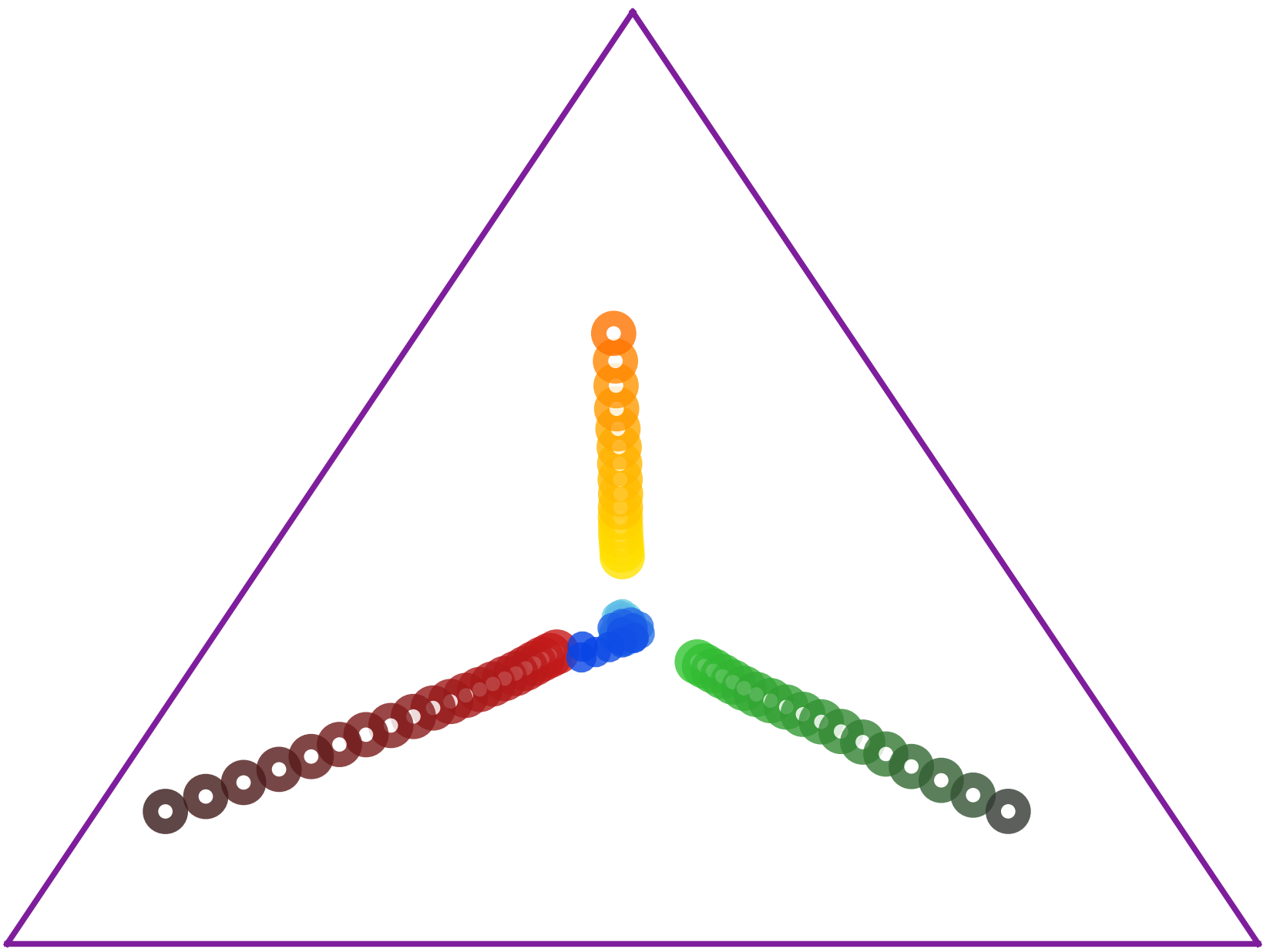}
\end{centering}
}
\subfloat[]{\begin{centering}
\includegraphics[width=0.7125\linewidth]{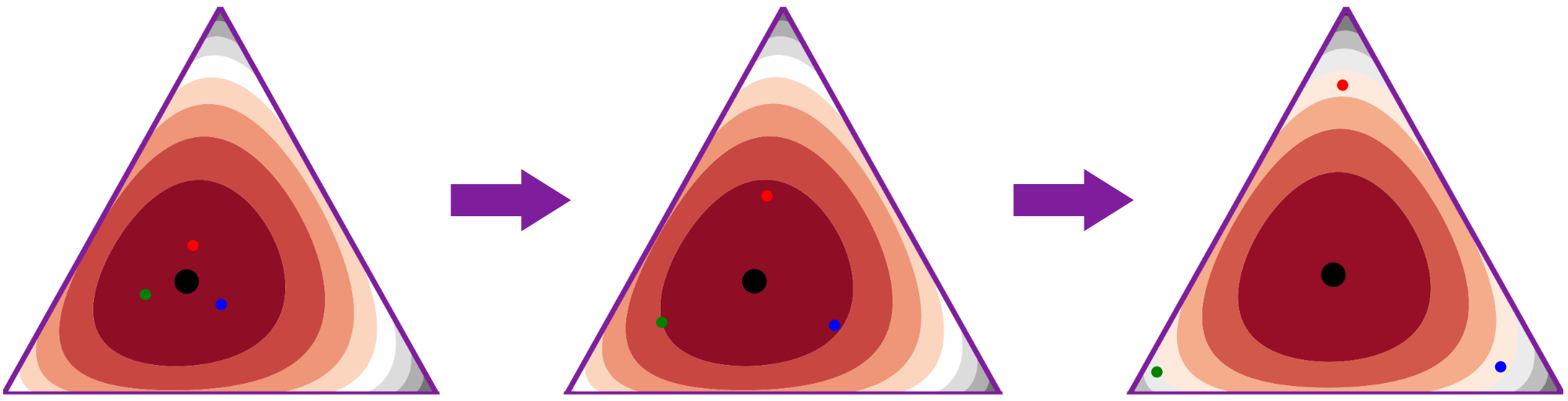}
\end{centering}
}
\caption{(a) Policy evolution trajectory. We initialize three policies here, denoted by red, yellow, and green circles on the simplex. The darker the color of the policy, the more iterations it has gone through, and the greater the diversity distance between this policy and other policies is. Moreover, the blue circles on the simplex denote the average state marginal distribution of policies $\rho(s)$. (b) Policy evolution process. We initialize three policies here as well, denoted by red, green, and blue dots on the simplex. The black dot denotes the average state marginal distribution of policies $\rho(s)$. Moreover, the contour lines in the figure correspond to the diversity metric $I(s;z)$.}
\label{fig:exp1}
\end{figure}

\section{Experiments}
This section presents some experimental results about diversity algorithms. Firstly, from an intuitive geometric perspective, we demonstrate the process of policy evolution in the diversity algorithm. Then we compare the three policy selection methods mentioned in Section~\ref{subsection: unified} by experiments, which illustrates the high efficiency of {\bf bandit selection}. 

\begin{figure}[t]
\center
\subfloat[]{\begin{centering}
\includegraphics[width=0.95\linewidth]{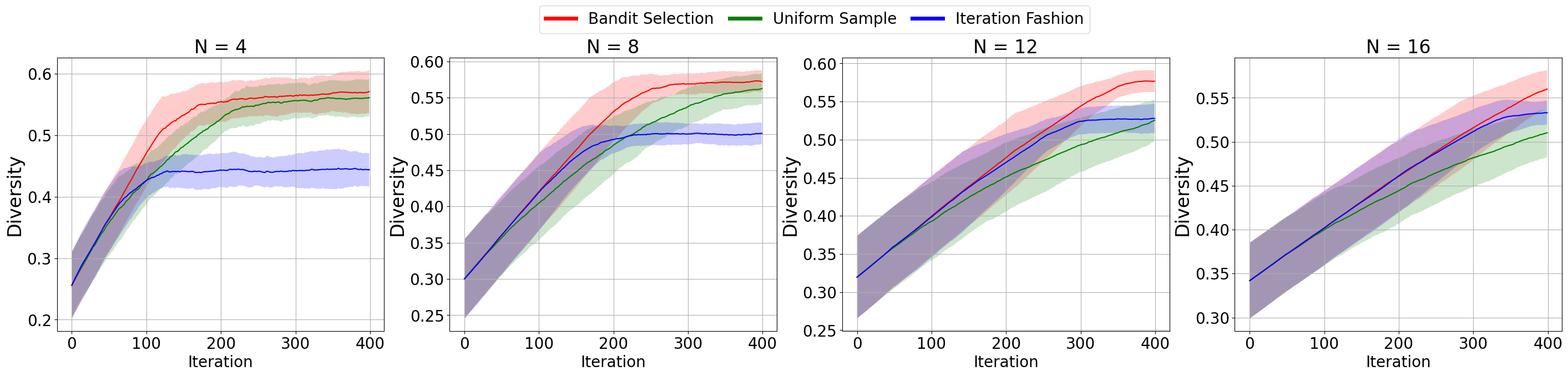}
\end{centering}
}

\subfloat[]{\begin{centering}
\includegraphics[width=0.95\linewidth]{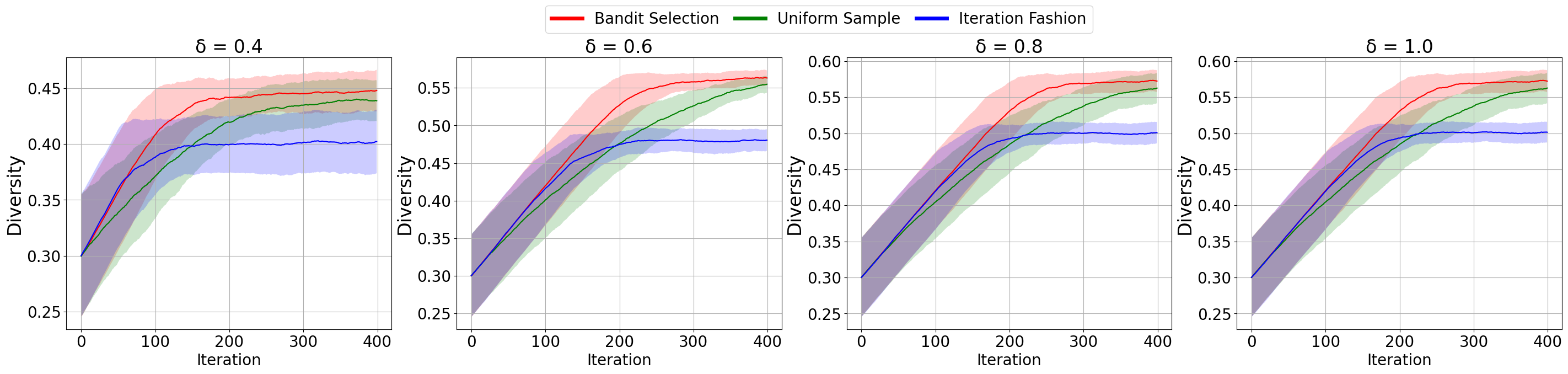}
\end{centering}
}

\caption{Comparison of different policy selection methods. 
(a) Training curves for different numbers of policies with a fixed $\delta$-\emph{target} where $\delta = 0.8$. (b) Training curves for different $\delta$-\emph{target} with a fixed number of policies where $N = 8$.}
\label{fig: Policy Selection}
\end{figure}

\subsection{A Geometric Perspective on Policy Evolution}

To visualize the policy evolution process, we use DIAYN~\cite{eysenbach2018diversity} as our diversity algorithm and construct a simple 3-state MDP~\cite{eysenbach2021information} to conduct the experiment. The set of feasible state marginal distributions is described by a triangle $[(1,0,0),(0,1,0),(0,0,1)]$ in $\mathbb{R}^3$. And we use state occupancy measure $\rho^{\pi_i}(s)$ to represent policy $\pi^i$. Moreover, we project the state occupancy measure onto a two-dimensional simplex for visualization. 

Let $\rho(s)$ be the average state marginal distribution of all policies. Figure~\ref{fig:exp1}(a) shows policy evolution during training. Initially, the state occupancy measures of different policies are similar. However, as training progresses, the policies spread out, indicating increased diversity. Figure~\ref{fig:exp1}(a) highlights that diversity~\cite{eysenbach2021information} ensures distinct state occupancy measures among policies.

We use $I(\cdot;\cdot)$ to denote mutual information. The diversity metric in unsupervised skill discovery algorithms is based on the mutual information of states and latent
variable $z$. Furthermore, the mutual information can be viewed as the average divergence between each policy’s state distribution $\rho(s|z)$ and the average state distribution $\rho(s)$~\cite{eysenbach2021information}:

\begin{equation}\label{eq6}
I(s;z) = {\bf E}_{p(z)}[D_{KL}(\rho(s|z) \parallel  \rho(s))].
\end{equation}

Figure~\ref{fig:exp1}(b) shows the policy evolution process and the diversity metric $I(s;z)$. We find that the diversity metric increased gradually during the training process, which is in line with our expectation.

\subsection{Policy Selection Ablation}
We continue to use 3-state MDP~\cite{eysenbach2021information} as the experimental environment. Whereas, in order to get closer to the complicated practical environment, we set specific $\delta$-\emph{target} and increased the number of policies. Moreover, when a policy that hasn't met the diversity requirement is chosen to update, we will receive a reward $r=1$, otherwise, we will receive a reward $r=0$. We use $I(s;z)$ as the diversity metric and use LinUCB\cite{li2010contextual} as our contextual bandit algorithm.  

Figure~\ref{fig: Policy Selection} shows the training curves under different numbers of policies and different $\delta$-\emph{target} over six random seeds. The results show that {\bf bandit selection} not only always reaches the convergence fastest, but also achieves the highest overall diversity metric of the population when it converges. We now empirically analyze the reasons for this result: \\
{\bf Drawbacks of uniform sample.} In many experiments, we observe that {\bf uniform sample} has similar final performance to {\bf bandit selection}, but significantly slower convergence. This is because after several iterations, some policies become distinct enough to prioritize updating other policies. However, {\bf uniform sample} treats all policies equally, resulting in slow convergence. \\
{\bf Drawbacks of iteration fashion.} In experiments, the {\bf iteration fashion} converges quickly but has lower final performance than the other two methods. It's greatly affected by initialization. Each policy update depends on the previous one, so poor initialization can severely impact subsequent updates, damaging the overall training process. \\
{\bf Advantages of bandit selection.} Considering historical rewards and balancing exploitation and exploration, {\bf bandit selection} quickly determines if a policy is different enough to adjust the sample's probability distribution. Unlike {\bf iteration fashion}, all policies can be selected for an update in a single iteration, making {\bf bandit selection} not limited by policy initialization.

\section{Conclusion}
In this paper, we compare existing diversity algorithms, provide a unified diversity reinforcement learning framework, and investigate the convergence of training diverse policies. Moreover, we propose {\bf bandit selection} under our proposed framework, and present the \emph{regret} bound for it. Empirical results
indicate that {\bf bandit selection} achieves the highest diversity score with the fastest convergence speed compared to baseline methods. We also provide a geometric perspective on policy evolution through experiments. In the future, we will focus on the comparison and theoretical analysis of different reward calculation methods. And we will continually explore the application of diversity RL algorithms in more real-world decision-making tasks.
%
%
%
\bibliographystyle{splncs04}
\bibliography{reference}
%




\end{document}